\documentclass[runningheads]{llncs}
\usepackage{graphicx}
\usepackage{amsmath}
\usepackage{amssymb}
\usepackage{multirow}
\usepackage{subcaption}
\usepackage{booktabs}
\usepackage{hyperref}
\usepackage[font=small,skip=5pt]{caption}
\begin{document}
\title{Representation Disentanglement for Multi-modal Brain MRI Analysis}
\titlerunning{Representation Disentanglement for Multi-modal Brain MRI Analysis}

\author{Jiahong Ouyang\inst{1 [0000-0002-0434-5757]} \and
Ehsan Adeli\inst{1 [0000-0002-0579-7763]} \and \\
Kilian M Pohl\inst{1,2 [0000-0001-5416-5159]} \and
Qingyu Zhao\inst{1 [0000-0002-9694-6022]} \and \\
Greg Zaharchuk\inst{1 [0000-0001-5781-8848]}}
\authorrunning{J. Ouyang et al.}
%\institute{Stanford University, Stanford CA 94305, USA \and SRI International, Menlo Park CA, 94025, USA\\}
\institute{$^1$Stanford University, Stanford, CA \quad $^2$SRI International, Menlo Park, CA\\ {\small \texttt{\{jiahongo,eadeli,kilian.pohl,qingyuz,gregz\}@stanford.edu}}}

% \author{Anonymous}
%
\maketitle              % typeset the header of the contribution
\begin{abstract}
Multi-modal MRIs are widely used in neuroimaging applications since different MR sequences provide complementary information about brain structures. Recent works have suggested that multi-modal deep learning analysis can benefit from explicitly disentangling anatomical (shape) and modality (appearance) information into separate image presentations. In this work, we challenge mainstream strategies by showing that they do not naturally lead to representation disentanglement both in theory and in practice. To address this issue, we propose a margin loss that regularizes the similarity in relationships of the representations across subjects and modalities. To enable robust training, we further use a conditional convolution to design a single model for encoding images of all modalities. Lastly, we propose a fusion function to combine the disentangled anatomical representations as a set of modality-invariant features for downstream tasks. We evaluate the proposed method on three multi-modal neuroimaging datasets. Experiments show that our proposed method can achieve superior disentangled representations compared to existing disentanglement strategies. Results also indicate that the fused anatomical representation has potential in the downstream task of zero-dose PET reconstruction and brain tumor segmentation.
The code is available at {\small \url{https://github.com/ouyangjiahong/representation-disentanglement}}.
\end{abstract}

\section{Introduction}
% use of multi-modality images 
Multi-modal MRIs using different pulse sequences (e.g., T1-weighted and T2 Fluid Attenuated Inversion Recovery) are widely used to probe complementary and mutually informative aspects of the brain structure, thereby playing a pivotal role in improving the understanding of neurodevelopment across the life span and diagnosis of neuropsychiatric disorders \cite{shen2020multi}. 
%Each modality provides unique information about the tissue, while depicting the similar anatomical structure of the scanned subject.
However, 
%the rich information encoded in the multi-modal imaging data complicates the design of image-based learning models, as they often assume the distribution of the input data lies in a single domain, which does not hold for multi-modal MRIs. Moreover, 
compared to uni-modal image analysis, models that operate on multi-modal data are more likely to encounter the issue of incomplete inputs (some cases have missing modalities) due to data corruption, when applied to larger MRI datasets \cite{lee2019collagan}.

To tackle these challenges, recent works \cite{shen2020multi,yang2020cross} have suggested to explicitly disentangle anatomical and modality-specific information from multi-modal MRIs. Specifically, each image is encoded into two representations: an \textit{anatomical representation} that encodes the morphological shape of brain anatomies and is mostly shared across all modalities of the same subject, and a \textit{modality representation} that encodes image appearance information specific to the modality. Such disentanglement is typically derived based on \textit{cross-reconstruction} \cite{lee2018diverse}, i.e., by examining the quality of images synthesized from anatomical and modality representations from mixed sources. The resulting disentangled representations are shown to be useful for downstream tasks including cross-modality deformable registration \cite{qin2019unsupervised}, multi-modal segmentation \cite{yang2020cross}, image harmonization \cite{dewey2020disentangled}, multi-domain image synthesis, and imputation of missing modalities \cite{shen2020multi}. 
%Another potential application is to handle incomplete multi-modal inputs \cite{shen2020multi}. In practice, some modalities can be missing due to different imaging protocols and data corruption, etc. which makes training and testing models challenging. The fully-disentangled anatomical representations can potentially solve this. 

All the above studies focused on evaluating the results of the downstream tasks. It remains \textit{unclear} whether the learned representations are truly disentangled or not. In this work, we show that the cross-reconstruction strategies can easily lead to information leakage between representations, i.e., representations are still partly coupled after disentanglement. To address this issue, we propose a margin loss that regularizes the within-subject across-modality similarity between representations with respect to the across-subject within-modality similarity. Such regularization encourages the anatomical and modality information to fully disentangle in the representation space. Further, to obtain a robust training scheme, we use a modified conditional convolution to combine separate encoders associated with the modalities into a single coherent model. Lastly, we introduce a fusion function to combine the disentangled anatomical representations as a set of modality-invariant features, which can be used to solve various downstream tasks. We evaluate our method on three multi-modal neuroimaging datasets, including T1- and T2-weighted MRIs of 692 adolescents from the National Consortium on Alcohol and Neurodevelopment in Adolescence (NCANDA) \cite{zhao2020association}, T1-weighted and T2-FLAIR MRIs of 173 adults for zero-dose PET reconstruction, and multi-modal MRIs (T1, post-contrast T1, T2, and T2 Fluid Attenuated Inversion Recovery) from 369 subjects of the BraTS 2020 dataset \cite{menze2014multimodal}. Results indicate that our method achieves better disentanglement between anatomical and modality representations compared to several baseline methods. The fused modality-invariant representation shows potential in the downstream task of PET reconstruction and brain tumor segmentation (BraTS).

\section{Related Works}
Representation disentanglement is an active topic in image-to-image translation tasks lately \cite{lee2018diverse}. The goal of these tasks is to disentangle the content (e.g., anatomical information) and style (e.g., modality, texture, appearance) information from an image so that images of the same content can be translated between different styles. The disentanglement is learned by optimizing a cross reconstruction loss on synthesized images, with content and style sampled from different training images \cite{huang2018multimodal,chartsias2019disentangle}. 
%Other strategies are based on regularizing the similarity on the anatomical representations \cite{dewey2020disentangled}, and latent consistency \cite{huang2018multimodal,chartsias2019disentangle}. 
A major issue is that these methods do not explicitly enforce the disentanglement, and hence the learned representations still suffer from information leakage.
%However, the major issue is that these methods do not explicitly guarantee the quality of disentanglement as each domain has its domain-specific decoder. Therefore, regardless of being disentangled or not, the learned content representations should be able to produce satisfying results for downstream tasks as long as they carry task-related information from the input.

Based on this observation, methods based on adversarial training \cite{lee2018diverse,liu2018multi,denton2017unsupervised,benaim2019domain} further regularize the content representations to be independent of the source style domain.
%apply adversarial training to further regularize the content representations in the same style domain.
For example, DRIT \cite{lee2018diverse} couples adversarial training with a cross-cycle consistency loss to achieve between-domain translation based on unpaired data. MTAN \cite{liu2018multi} uses a multi-class adversarial loss for the style labels. DRNet \cite{denton2017unsupervised} leverages the adversarial loss to disentangle the stationary and temporal components. Sagie et al. \cite{benaim2019domain} proposed a zero loss to force the style encoder to capture information relevant to the specific domain. 
%Similarly, the adversarial training strategy is applied to medical applications to handle multi-modal data analysis, such as lung CT or brain MR registration \cite{qin2019unsupervised}, brain and prostate MR data imputation \cite{shen2020multi}, and liver segmentation based on CT and MR \cite{yang2020cross}. 
However, adversarial training can be unstable and easily stuck in a local optimum. %\cite{salimans2016improved}. 
In addition, only one of the two representations (usually the content representation) can be adversarially regularized, which can easily cause information leakage into the other representation. As an extreme scenario, the style representation can contain all the information about the input, while the content representation may carry little or random information, just enough to fool the discriminator. 

Lastly, a common issue of the above methods is that they utilize a separate decoder for each domain. It means that regardless of being disentangled or not, the learned content representations can always produce satisfying results for domain translation or downstream tasks as long as they carry task-related information from the input. In conclusion, in the absence of visualization or evaluation of the disentangled representations, it is unclear if the representations learned by the above methods are still coupled or not. 

% InfoGAN \cite{chen2016infogan} and $\beta$-VAE \cite{higgins2016beta} has been proposed to disentangle salient attributes in an unsupervised manner.
% fetal ultrasound synthesis \cite{meng2019representation},
% ,yang2019unsupervised

% not sure if it's appropriate in this section. and it might be a bit rambling
%\noindent \textbf{Discussion}  For the second category, 

\section{Proposed Method}
% a graph to show two-stage process
To address this ambiguity, we first introduce a robust model for disentangling the anatomical and modality representations for multi-modal MRI based on image-to-image translation in Section 3.1. 
%Our model operates in a self-supervised manner and does not require any external label information for training. 
Next, we introduce a strategy for fusing the disentangled anatomical representations from all available modalities of a subject into a modality-invariant representation, which can be used as the input for any downstream model.

% TODO: will change it to another figure that better shows the loss and the flow
\begin{figure}[t]
\centering
\includegraphics[width=0.90\textwidth,trim=0 10 0 0,clip]{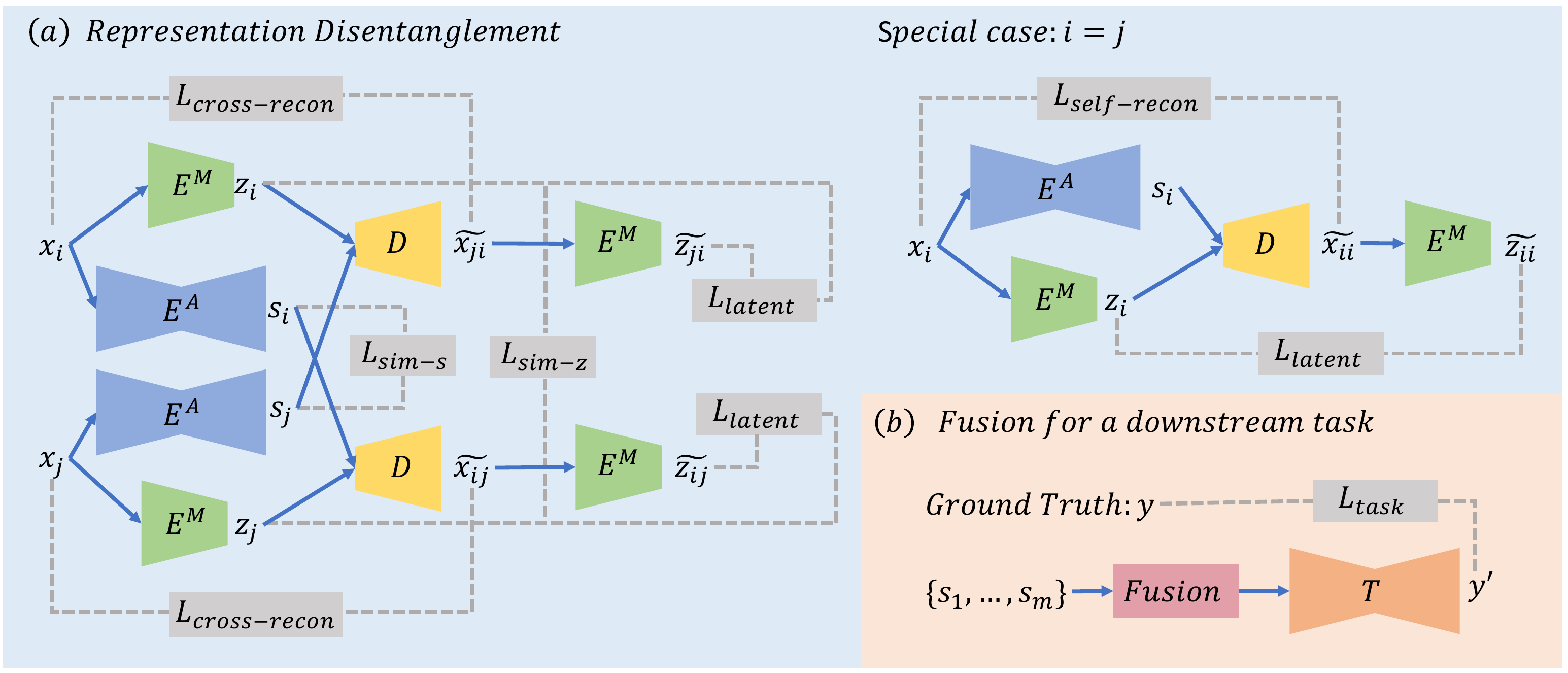}
% \vspace{-25pt}
\caption{Overview: (a) An image $x_i$ is disentangled into an anatomical representation $s_i$ and a modality representation $z_i$ by $E_A$ and $E_M$. The decoder $D$ reconstructs the input from the two representations. These networks are trained by the reconstruction and latent consistency losses. We propose to add a similarity regularization $L_{sim}$ that models the relationships between the representations from different images; (b) The disentangled anatomical representations of a subject are fused into one modality-invariant encoding that can be used as an input to a downstream model $T$.} 
\label{fig:overview}
\end{figure}

\subsection{Representation disentanglement by image-to-image translation}

We assume each subject in the training set has MRIs of $m$ modalities (sequences) and let $x_i \in \mathcal{X}_i$ denote the input image of the i-th modality. As shown in Fig. 1a, we aim to disentangle $x_i$ into an anatomical representation $s_i$ by an anatomical encoder $s_i=E_i^A(x_i)$ and a modality representation $z_i$ by a modality encoder $z_i=E_i^M(x_i)$. We assume that the anatomical representation $s_i$ encodes the morphological information of brain structures that is mostly impartial to the imaging modality, while $z_i$ provides image appearance information specific to a modality. The decoder $D$ then reconstructs $x_i$ from a pair of anatomical and modality representations. %It can be the corresponding representations $s_i$ and $z_i$ or the mix-and-match representations $s_j$ and $z_i$. 
Prior works \cite{chartsias2019disentangle} have suggested that such disentangled representations can be learned by optimizing the \textit{self-reconstruction} and \textit{cross-reconstruction} losses; Given a pair of $s_i$ and $z_j$ derived from images of any two modalities, $D$ is supposed to synthesize an image that is similar to the input image $x_j \in \mathcal{X}_j$, whose synthesized domain corresponds to the j-th modality.
\vspace{-8pt}
\begin{align*}
    L_{self-recon} &= \frac{1}{m}\sum_{i=1}^{m} \mathbf{E}_{x_i \sim \mathcal{X}_i} [\parallel \widetilde{x_{ii}} - x_i \parallel _1], \\
    L_{cross-recon} &= \frac{\lambda_c}{m^2-m} \sum_{i=1}^{m} \sum_{j=1, j\neq i}^m \mathbf{E}_{x_i \sim \mathcal{X}_i, x_j \sim \mathcal{X}_j} [\parallel \widetilde{x_{ij}} - x_j \parallel _1], 
\end{align*}
where $\widetilde{x_{ij}} = D(E_i^A(x_i), E_j^M(x_j))$. %Note, the expectation term corresponds to the cross-reconstruction loss when $i \neq j$ and the self-reconstruction loss when $i=j$ (Fig. 1a). 
In addition to these reconstruction losses, another loss function commonly used for training image-to-image translation \cite{qin2019unsupervised} is the \textit{latent consistency} loss, which encourages the representations derived from raw inputs to be similar to the ones from the synthesized images. 

\vspace{-8pt}
\begin{equation}
    L_{latent} = \frac{\lambda_l}{m^2}\sum_{i=1}^m \sum_{j=1}^m \mathbf{E}_{x_i \sim \mathcal{X}_i, x_j \sim \mathcal{X}_j} [\parallel \widetilde{z_{ji}} - z_i \parallel _1],
\end{equation}
where $\widetilde{z_{ji}} = E_i^M(D(E_j^A(x_j), E_i^M(x_i)))$ is the modality representation derived from a synthesized image.

% \begin{equation}
%     L_{latent} = \sum_{i=1}^m \mathbf{E}_{x_i \sim \mathcal{X}_i} [\parallel E_i^M(D(E_i^A(x_i), E_i^M(x_i))) - E_i^M(x_i) \parallel _1] \\
% \end{equation}

%\begin{gather*}
%    L_{self-recon} = \sum_{i=1}^m \mathbf{E}_{x_i \sim \mathcal{X}_i} [\parallel D(E_i^A(x_i), E_i^M(x_i)) - x_i \parallel _1] \\
%    L_{cross-recon} = \sum_{i=1}^{m-1} \sum_{j=i+1}^m \mathbf{E}_{x_i \sim \mathcal{X}_i, x_j \sim \mathcal{X}_j} [\parallel D(E_i^A(x_i), E_j^M(x_j)) - x_j \parallel _1] \\
    % L_{latent-s} = \sum_{i=1}^m \mathbf{E}_{s_i \sim \mathcal{S}} [\parallel E_i^A(D(E_i^A(x_i), E_i^M(x_i))) - E_i^A(x_i) \parallel _1] \\
%    L_{latent} = \sum_{i=1}^m \mathbf{E}_{x_i \sim \mathcal{X}_i} [\parallel E_i^M(D(E_i^A(x_i), E_i^M(x_i))) - E_i^M(x_i) \parallel _1] \\
%\end{gather*}

% similarity loss
\noindent\textbf{Enforcing Disentanglement by Similarity Regularization.} Although prior works have leveraged the above concept of disentanglement for several multi-modal downstream tasks, there is no theoretical guarantee that the cross reconstruction can encourage the encoder to disentangle anatomical and modality representations. In fact, information can freely leak between representations. As a naive example, both $s_i$ and $z_i$ can be an exact copy of $x_i$ so that the decoder $D$ can easily reconstruct the input.   

We resolve this problem by exploring the similarity relationships between representations. As the brain's morphological shape is highly heterogeneous across subjects, we assume the anatomical representations $s$ from the same subject but different modalities should be more similar than those from the same modality but different subjects. 
%from all input modalities should locate closely in $\mathcal{S}$,
Note, $s_i$ of the same subject are not necessary to be exactly the same, as multi-modal imaging is designed to capture distinct characteristics of brain anatomies. For instance, the brain tumor itself is more visible on T1-weighted MR with contrast (T1c) compared to T1 without contrast due to the injected contrast medium. On the other hand, modality representations $z$ from the same modality but different subjects should be more similar than those from the same subject but different modalities.
%should describe similar information about the tumor, but T1c may capture more . 
%However, it can not ensure the disentangled representations. e.g. $s$ may still carry some modality information. 
We propose to model such relationships using a similarity loss inspired by the margin-based hinge loss \cite{frome2013devise}. 
\vspace{-4pt}
\begin{align}
     L_{sim} = & \frac{\lambda_s}{m^2} \sum_{i=1}^{m} \sum_{j=1}^m \mathbf{E}[\text{max}(0, \alpha_s - \text{cos}(f(s_i^p), f(s_j^p)) + \text{cos}(f(s_i^p), f(s_i^q)))] + \nonumber\\
    & \frac{\lambda_z}{m^2} \sum_{i=1}^{m} \sum_{j=1}^m \mathbf{E}[\text{max}(0, \alpha_z - \text{cos}(z_i^p, z_i^q) + \text{cos}(z_i^p, z_j^p))]
    \label{eq:sim}
    % L_{sim} = &\lambda_s \frac{2}{m^2-m} \sum_{i=1}^{m-1} \sum_{j=i+1}^m \mathbf{E}[\text{max}(0, \alpha_s - \text{cos}(f(s_i^p), f(s_j^p)) + \text{cos}(f(s_i^p), f(s_i^q)))] + \nonumber\\
    % &\lambda_z \frac{2}{m^2-m} \sum_{i=1}^{m-1} \sum_{j=i+1}^m \mathbf{E}[\text{max}(0, \alpha_z - \text{cos}(z_i^p, z_i^q) + \text{cos}(z_i^p, z_j^p))]
    % \label{eq:sim}
\end{align}
where $p$ and $q$ correspond to a pair of subjects randomly sampled in a mini-batch, $\text{cos}(\cdot,\cdot)$ denotes the cosine distance between two vectors, and $f$ denotes a MaxPooling and flattening operation. Unlike the L$_2$-based similarity loss, Eq. \eqref{eq:sim} encourages the within-subject and across-subject distances to differ by the margins $\alpha_s$ and $\alpha_z$ and thereby avoids deriving identical representations.

% data imputation for incomplete input
%\noindent \textbf{Imputation for incomplete inputs} As described above, the model is a discriminative model that cannot sample the synthesized images. To achieve image-to-image translation from $\mathcal{X}_i$ to $\mathcal{X}_j$ in testing phase, for sample $p$, it is synthesized by $D(s_i^p, h(z_j))$, where $h(z_j)$ can be the average of $z_j$ for the corhort or the $z_j$ from nearest neighbour of sample $p$ in $\mathcal{S}$.

% cond-conv for multi-domain translation 
\noindent \textbf{Conditional convolution.} Another drawback of traditional multi-modal image translation methods is that each modality is associated with a pair of anatomical and modality encoders that are independently learned. However, these encoding tasks are highly dependent across modalities. Hence, each convolutional operation at a certain layer should function similarly across networks. To enable robust multi-modal translation, we couple the set of independent anatomical encoders $\{E_i^A(x_i)\}$ into one coherent encoder model $E^A(x;i)$ and likewise couple all modality encoders $\{E_i^M(x_i)\}$ into $E^M(x;i)$ with $i$ being an additional input to the model. We construct these two unified models using Conditional Convolution (CondConv) \cite{yang2019condconv} as the fundamental building blocks. As inspired by \cite{yang2019condconv}, parameters of a convolutional kernel is conditioned on the input modality $i$ using a mixture-of-experts model $CondConv(x;i) = \sigma ((\beta_1^i \cdot W_1 + ... + \beta_n^i \cdot W_n) \circledast x)$,
%We modify the convolution kernels to conditioned on the input modality, which can be parameterized by:
where $\sigma(\cdot)$ is the sigmoid activation function, $\circledast$ denotes regular convolution, $\{W_1,...,W_n\}$ are the learnable kernels associated with $n$ experts, and $\{\beta^i_1,...\beta^i_n\}$ are the modality-specific mixture weights. As such, the convolutional kernel exhibits correlated behavior across modalities as the $n$ experts are jointly trained on data of all modalities.
%$\beta = sigmoid(W_{beta} \cdot i)$ is a vector of modality-dependent weights. 

\subsection{Fusing disentangled representations for downstream tasks}
% way to fuse features
As shown in Fig. 1b, after obtaining the disentangled representations, the anatomical representations from all available modalities of a subject are fused into one fixed-size encoding as the input for a downstream model $T$, which can be any state-of-the-art model for the downstream task. Note that the fusion function here should pool features of a various number of channels to a fixed number of channels. Let $s$ be the concatenation of anatomical representations from the available modalities $Concat(s_i, ..., s_j)$. Then the fusion is the concatenation of several pooling functions: $Concat(MaxPool(s), MeanPool(s), MinPool(s)).$
% Here, we define the fusion as the concatenation of the output of several basic pooling functions: 
% \begin{equation*}
%     Fusion(s) = Concat(MaxPool(s), MeanPool(s), MinPool(s)).
% \end{equation*}
With this fusion operation, one can use two strategies to train the downstream model $T$. We can either solely train $T$ based on the frozen $s$ derived by the self-supervised learning of the encoders (Section 3.1), or fine-tune the encoders jointly with the downstream task model $T$. Though the joint training can potentially result in representations that better suit the downstream task, we confine our analysis to the first strategy (fixing encoders) in this work to emphasize the impact of representation disentanglement.

\section{Experiments}
We first describe the dataset and the experimental settings in Section 4.1 and 4.2. We then show in Section 4.3 that our proposed approach (Section 3.1) can effectively disentangle anatomical and modality representations on three neuroimaging datasets. We further show in Section 4.4 that the disentangled representations in combination with the fusion strategy (Section 3.2) can alleviate the missing modality problem in two downstream tasks.  
\subsection{Datasets}
\textbf{ZeroDose} The dataset comprised brain FDG-PET and two MR modalities (T1-weighted and T2 FLAIR) from 171 subjects with multiple diagnosis types including tumor, epilepsy, and dementia. The FLAIR and PET images were first registered to T1, and then all modalities were normalized to a standard template and resized to $192 \times 160$ in the axial plane. Intensities in the brain region of each image were converted to z-scores. The top and bottom 20 slices in each image were omitted from analysis. Each 3 adjacent axial slices were converted to a 3-channel image as the input to the encoder models. Random flipping of brain hemispheres was used as augmentation during training. Five-fold cross-validation was conducted with 10\% training cases used for validation. The downstream task was zero-dose PET reconstruction, i.e., to synthesize high quality FDG-PET from multi-modal MRIs. This is useful in practice as the injected radiotracer in current PET imaging protocols can lead to the risk of primary or secondary cancer in scanned subjects. Moreover, PET is more expensive than MRI and not offered in the majority of medical centers worldwide. %for zero-dose PET reconstruction, meaning that using MR images that of no radiotracer to reconstruct PET images.  The downstream task is to synthesize high quality FDG-PET with multi-modal MR contrasts. 

\noindent\textbf{NCANDA} Based on the public data release\footnote[1]{NCANDA\_PUBLIC\_4Y\_STRUCTURAL\_V01 (DOI: 10.7303/syn22216457)%; distributed to the public according to the NCANDA Data Distribution agreement \cite{zhao2020association} \url{https://www.niaaa.nih.gov/ncanda-data-distribution-agreement}
; collection was supported by NIH grants AA021697, AA021695, AA021692, AA021696, AA021681, AA021690, and AA02169}, we used the T1 and T2 MRIs of 692 adolescents with no-to-low alcohol drinking from the NCANDA dataset \cite{zhao2020association}. The images were preprocessed by a pipeline \cite{zhao2019confounder} composed of denoising, bias field correction, skull stripping, aligning T1 and T2 to a template, and resizing to $192 \times 160$. Other settings were the same as the ZeroDose dataset. As this dataset only contained healthy individuals, we used it solely for evaluating the representation disentanglement based on the middle 40 slices in each image.

\noindent\textbf{BraTS} Multimodal Brain Tumor Segmentation Challenge 2020 \cite{menze2014multimodal} provides multi-modal brain MRI of 369 subjects with four modalities: T1, post-contrast T1 (T1Gd), T2, and T2-FLAIR (FLAIR). Three categories were labeled for brain tumor segmentation, i.e., Gd-enhancing tumor (ET), peritumoral edema (ED), and necrotic and non-enhancing tumor core (NCR/NET). We used the 55 middle axial slices and cropped the image size to $192 \times 160$. Other preprocessing steps and settings were kept the same.

\subsection{Experimental Settings}
\textbf{Implementation Details} The anatomical encoder $E^A$ was a U-Net type model. Let C$_k$ denote a Convolution-BatchNorm-ReLU block with $k$ filters ($4\times4$ spatial filters with stride 2), and CD$_k$ an Upsample-Convolution-BatchNorm-ReLU block. The architecture was designed as C$_{32}$-C$_{64}$-C$_{128}$-C$_{256}$-C$_{256}$-CD$_{256}$-CD$_{128}$-CD$_{64}$-CD$_{32}$. A convolution then mapped the resulting representations to 4 channels with softmax activation as the anatomical representation. The modality encoder $E^M$ consisted of 5 convolution layers of 3 x 3 filters and stride 2 with LeakyReLU of a 0.2 slope. Numbers of filters were 16-32-64-128-128. A fully connected layer mapped the resulting features to a 16-D representation. The decoder $D$ was based on SPADE \cite{park2019semantic} with the architecture used in \cite{chartsias2019disentangle}. The networks were trained for 50 epochs by the Adam optimizer with learning rate of $2 \times 10^{-4}$ and weight decay of $10^{-5}$. %Learning rate is reduced by 0.1 if the validation loss is not improved for 5 epochs. 
The regularization rates were set to %$\lambda_{self-recon}=1.0$, 
$\lambda_{c}=2.0$,
$\lambda_{l}=0.1$, $\lambda_{s}=10.0$, $\lambda_{z}=2.0$. The margins in the similarity loss were set to $\alpha_s = \alpha_z=0.1$. For the downstream model $T$ in the zero-dose PET reconstruction, a U-Net based model with attention modules \cite{ouyang2020zero} was adopted. The downstream model for BraTS brain tumor segmentation was the BraTS 2018 challenge's winner NVNet \cite{myronenko20183d}. 

%In training process, fused representations is computed by anatomical representations from all modality or one modality's representation is randomly dropped.

\noindent \textbf{Competing Methods} We first implemented the encoders using traditional convolution (training separate encoders), denoted as \textbf{Conv}. Based on this implementation, our disentanglement approach incorporating the similarity losses is denoted as \textbf{+Sim}. We then compared our approach with two types of methods. We term the first type \cite{huang2018multimodal,chartsias2019disentangle} that regularized the disentanglement merely using cross-reconstruction and latent consistency loss as \textbf{+NA}. The other type \cite{lee2018diverse,liu2018multi,denton2017unsupervised,benaim2019domain,qin2019unsupervised,shen2020multi,yang2020cross} that utilized adversarial training on the anatomical representations are termed as \textbf{+Adv}. To make fair comparison on the disentanglement strategies, all comparison methods used the same network structure for the encoder and decoder. Finally, we replaced \textbf{Conv} with \textbf{CondConv} (training a single encoder) to show the advantage of using conditional convolution.

\subsection{Evaluation on disentangled representation}

% figures of s
\begin{figure}[!t]
\centering
\includegraphics[width=\textwidth]{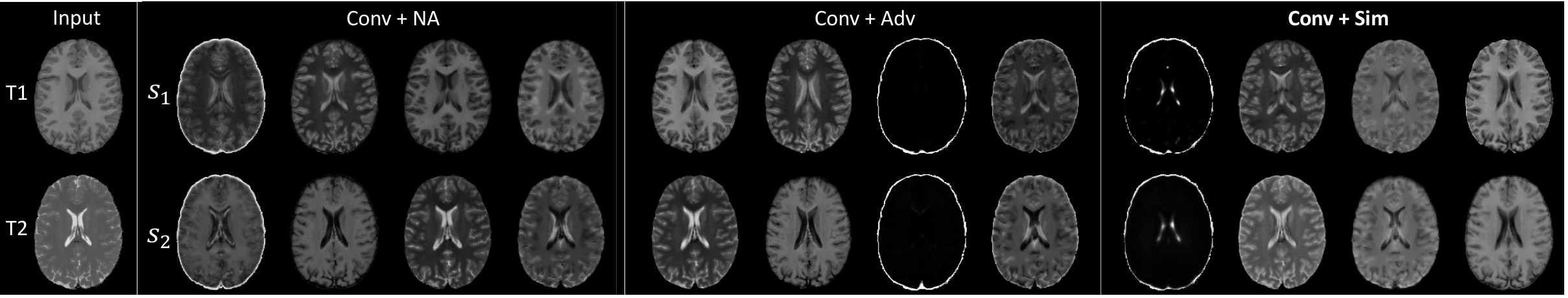}
% \vspace{-25pt}
\caption{Visualization of $s_1$ and $s_2$ of one NCANDA subject. Only our approach (+Sim) resulted in visually similar  anatomical representations from T1 and T2.} 
\label{fig:s-ncanda}
\end{figure}

% clustering of z (with missing)
\begin{figure}[!t]
\centering
\includegraphics[width=0.90\textwidth]{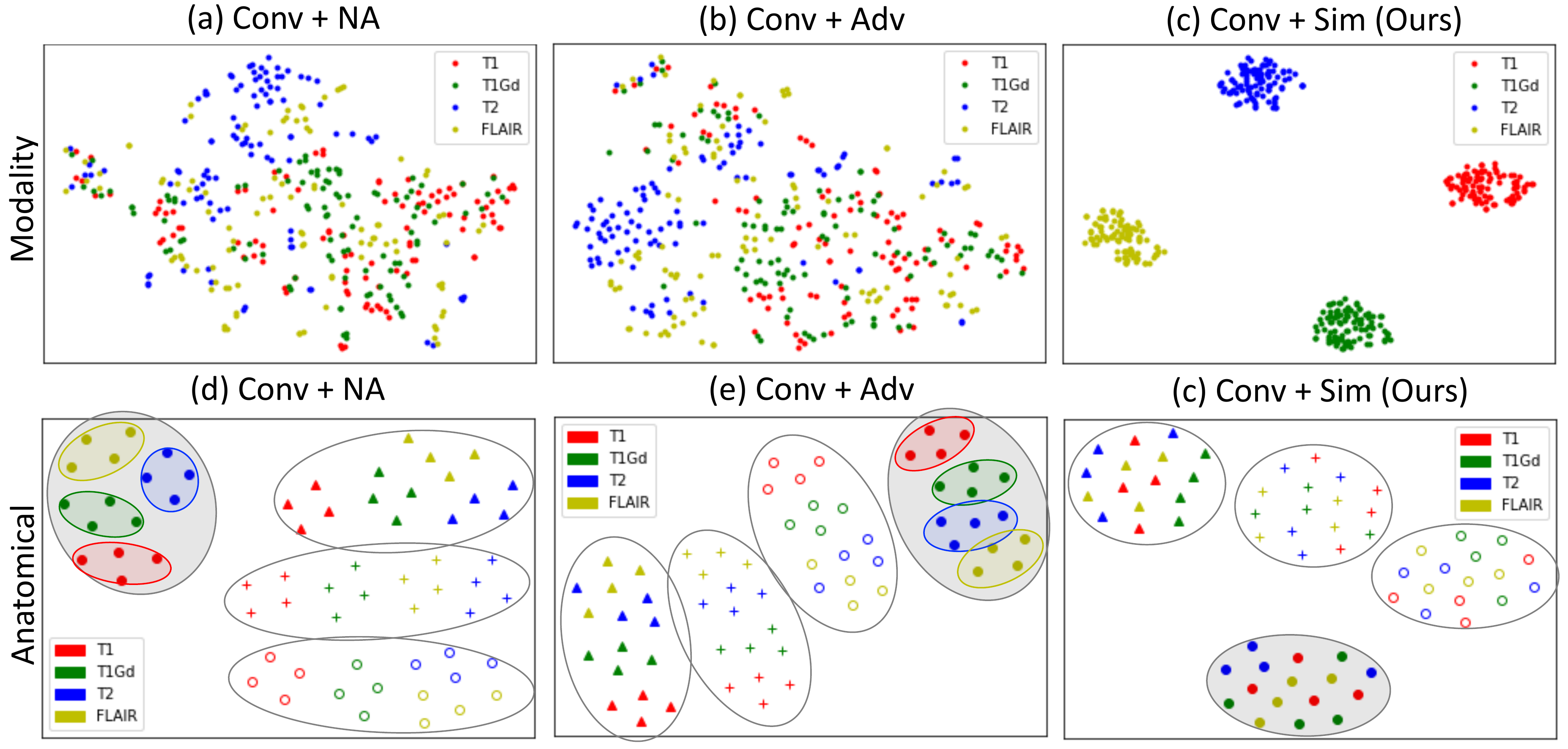}
% \vspace{-25pt}
%\caption{Visualization of modality representations $z$ from BraTS dataset.} 
% \includegraphics[width=\textwidth]{figure_s_BraTS.png}
% \vspace{-25pt}
\caption{t-SNE visualization of the $z$ space (a,b,c) and the $s$ space (d,e,f) for BraTS dataset in 2D spaces.
%(a,b,c) show the space of $z$ color-coded by modality.
Fully disentangled $z$ should cluster by modality (denoted by color); 
%(d,e,f) show the space of $s$ with each subject denoted by a different marker style. 
Fully disentangled $s$ should cluster by subjects (denoted by marker style) with no modality bias (sub-clusters by modality).} 
\label{fig:sz-brats}
\end{figure}

% clustering of z (with missing)
%\begin{figure}[!b]
%\centering
%\includegraphics[width=\textwidth]{figure_s_BraTS.png}
% \vspace{-25pt}
%\caption{Visualization of modality representations $s$ from BraTS dataset. Each %shape represents one subject.} 
%\label{fig:s-brats}
%\end{figure}

We first evaluated the methods on representation disentanglement and image cross reconstruction based on 5-fold cross-validation. We derived the anatomical and modality representations of the test subjects learned by the approaches. Fig. \ref{fig:s-ncanda} visualizes the 4-channel anatomical representations of one subject from NCANDA. We observe that $s_1$ and $s_2$ (extracted from T1 and T2) learned by the two baselines (Conv+NA and Conv+Adv) were \textit{substantially different}, indicating that they might still contain modality-specific information. On the other hand, our approach (Conv+Sim) produced visually more similar anatomical representations than the baselines. This indicates that the proposed similarity regularization can decouple the modality-specific appearance features from the structural information shared between modalities.

This result is also supported by the visualization of the learned representation spaces. As shown in Fig. \ref{fig:sz-brats}a-c, we randomly selected 200 modality representations in the test set of the BraTS dataset and projected them into a 2D space by t-SNE. Only our approach clearly separated the representations of different modalities into 4 distinct clusters (Fig. \ref{fig:sz-brats}c), which was in line with the regularization on $z$ in Eq. \eqref{eq:sim}. The clustering with respect to modalities was not evident for the projections of the baseline approaches (Fig. \ref{fig:sz-brats}a,b), indicating that complimentary information had leaked into the modalities representations. Moreover, the baseline approaches failed to disentangle T1 and T1Gd, two contrasts with high visual resemblance, as the red and blue dots were coupled in the representation space. Likewise, we visualized the space of anatomical representations in Fig. \ref{fig:sz-brats}d-f. We randomly selected 4 subjects in the BraTS test set and projected the pooled anatomical representation $f(s_i)$ of 4 consecutive slices into a 2D space. Now the 4 distinct clusters of our approach were defined with respect to subjects as opposed to modalities, and there was no apparent bias of modality in each subject's cluster (Fig. \ref{fig:sz-brats}f), indicating the representations solely encoded subject-specific yet modality-invariant information. The representation spaces learned by the two baselines contained both subject-specific anatomical and modality information (Fig. \ref{fig:sz-brats}d,e); that is, although the projections could be separated by subjects (black circles), each subject-specific cluster could be further stratified by modalities (color of the markers). 

% MR translation figure (w/o missing)
\begin{figure}[!t]
\centering
\includegraphics[width=0.95\textwidth]{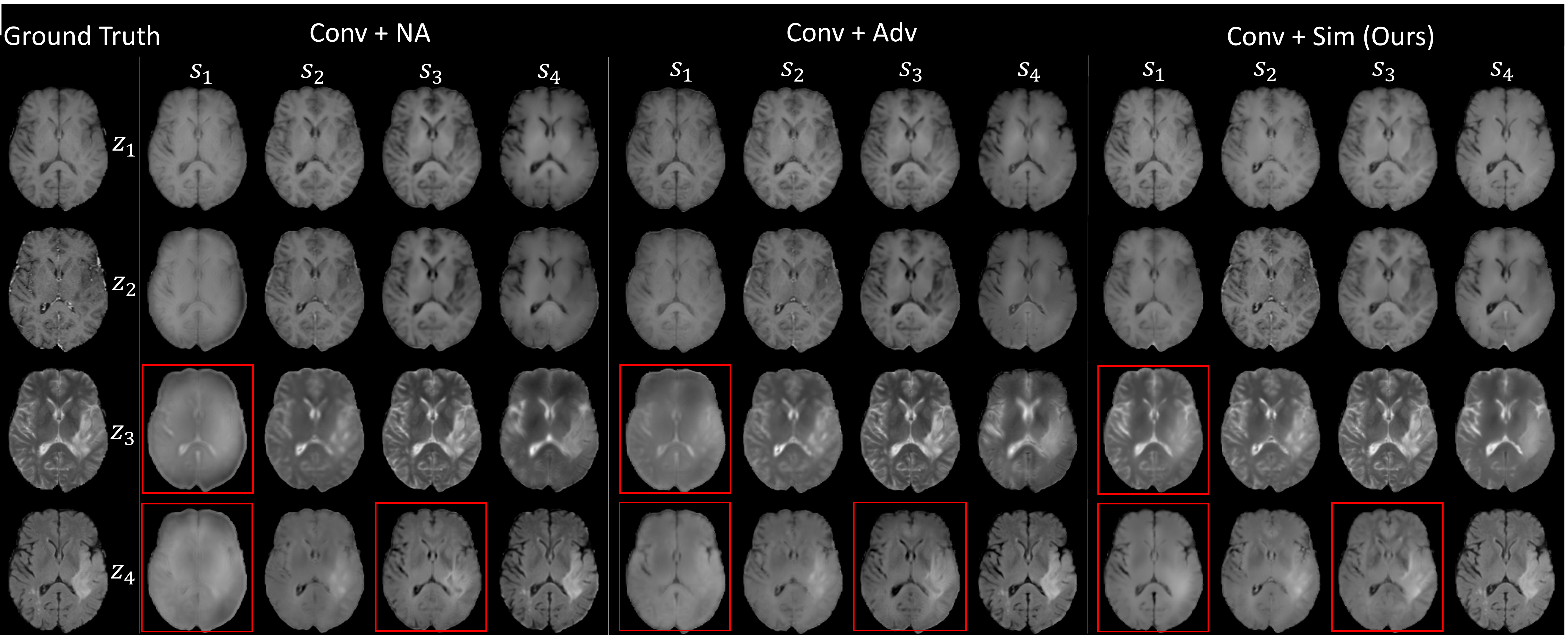}
\caption{Cross reconstruction results for one test subject from the BraTS dataset.} 
\label{fig:stage1-brats}
\end{figure}
\begin{table}[!t]
\tiny
% \scriptsize
\begin{subtable}{.5\linewidth}
\centering
\begin{tabular}{c|c|c}
\multicolumn{3}{c}{\small (a) ZeroDose}\\
    \toprule
    \multirow{2}{*}{Methods} & T1 & FLAIR \\
    \cline{2-3}
    % \multicolumn{2}{c}{\multirow{2}{*}{Multi-col-row}}&X\\
    % \multicolumn{2}{c}{}&X\\
    & PSNR/SSIM & PSNR/SSIM  \\
    \hline
    Conv+NA & 25.603/0.682 & 24.435/0.612\\
    Conv+Adv & 27.131/\textbf{0.742} & 25.846/0.674\\
    \textbf{Conv+Sim(Ours)} & 27.222/0.735 & 25.970/0.667\\
    \textbf{CondConv+Sim(Ours)} & \textbf{27.231}/\textbf{0.742} & \textbf{25.978}/\textbf{0.681}\\

    \bottomrule
\end{tabular}
%\caption{ZeroDose}
\label{tab:stage1-zerodose}
\end{subtable}%
\begin{subtable}{.5\linewidth}
\centering
\begin{tabular}{c|c|c}
\multicolumn{3}{c}{\small (b) NCANDA}\\
    \toprule
    \multirow{2}{*}{Methods} & T1 & T2 \\
    \cline{2-3}
    % \multicolumn{2}{c}{\multirow{2}{*}{Multi-col-row}}&X\\
    % \multicolumn{2}{c}{}&X\\
    & PSNR/SSIM & PSNR/SSIM \\
    \hline
    Conv+NA & 29.719/0.849 & 28.077/0.808\\
    Conv+Adv & \textbf{30.421}/\textbf{0.866} & 27.950/0.807\\
    \textbf{Conv+Sim(Ours)} & 30.266/0.863 & 28.367/0.825 \\
    \textbf{CondConv+Sim(Ours)} & 30.331/0.865 & \textbf{28.451}/\textbf{0.832}\\
    \bottomrule
\end{tabular}
%\caption{NCANDA}
\label{tab:stage1-ncanda}
\end{subtable}%
\vspace{5pt}
\begin{subtable}{\linewidth}
\centering
\begin{tabular}{c|c|c|c|c}
    \multicolumn{5}{c}{\small (c) BraTS}\\
    \toprule
    \multirow{2}{*}{Methods} & T1 & T1Gd & T2 & FLAIR\\
    \cline{2-5}
    & PSNR/SSIM & PSNR/SSIM & PSNR/SSIM & PSNR/SSIM\\
    \hline
    Conv+NA & 27.304/0.717 & 24.897/0.638 & 25.148/0.621 & 25.166/0.617 \\
    Conv+Adv & 27.485/0.717 & 25.385/0.656 & 25.951/0.658 & 26.135/0.642\\
    \textbf{Conv+Sim(Ours)} & 27.892/\textbf{0.756} & 26.114/0.723  & \textbf{26.479}/\textbf{0.744} & \textbf{26.588}/\textbf{0.692}\\
    \textbf{CondConv+Sim(Ours)} & \textbf{27.916}/0.752 & \textbf{26.221}/\textbf{0.731}  & 26.445/0.735 & 26.489/0.687\\
    \bottomrule
\end{tabular}
%\caption{BraTS}
\label{tab:stage1-brats}
\end{subtable}
\vspace{5pt}
\caption{5-fold cross-validation for quantitative cross-reconstruction evaluation.}
\label{tab:stage1}
\vspace{-8pt}
\end{table}

%We plotted   in a 2D space by t-SNE. Four modalities are represented by different colors, while different shapes label the 4 selected subjects. Though all three methods obtained clusters by subjects, only the proposed method has mixed anatomical representations, while others has sub-clusters by modality within each subject cluster. It demonstrates that the proposed similarity regularization can help to achieve better disentangled anatomical representations.

The improved disentanglement also resulted in better cross-reconstruction. Fig. \ref{fig:stage1-brats} shows the results of a test subject from the BraTS dataset. In each panel, $\widetilde{x_{ij}}$ is displayed on the $j^{th}$ row and $i^{th}$ column;
%Each row or column represent results using same modality representation or anatomical representation respectively. 
Diagonal images correspond to self-reconstruction and off-diagonal ones are cross-reconstruction. The proposed Conv+Sim achieved the best visual quality (accurate structural details), especially the FLAIR reconstruction highlighted in red boxes, where the tumor area was more precisely reconstructed. This improvement was quantitatively supported by the higher similarity between ground-truth and synthesized images in terms of peak-signal-noise ratio (PSNR) and structural similarity index (SSIM) (Table \ref{tab:stage1}). 
For each image of a specific modality, we synthesized it based on its own modality representation and the anatomical representation from another modality (For each image in the BraTS dataset, we computed the average metrics over all three cross reconstruction results). 
According to Table \ref{tab:stage1}, Conv+NA achieved the lowest reconstruction quality for both SSIM and PSNR on all three datasets. The quality improved when adding adversarial training on the anatomical representations (Conv+Adv) but was still lower than the two implementations with the proposed similarity loss (except for T1 reconstruction in NCANDA). %This suggests that the proposed similarity regularization (+Sim) is helpful in extracting information for cross reconstruction. 
Of the two models with +Sim, CondConv+Sim recorded better performance on the ZeroDose and NCANDA datasets. This indicates that CondConv enabled more stable reconstruction results by coupling highly dependent modality-specific encoders. However, on the BraTS dataset, Conv+Sim achieved better cross-reconstruction for T2 and FLAIR. The reason could be that CondConv shared anatomical and modality encoders across modalities at the expense of model capacity, especially when more than two modalities were involved. It is also worth mentioning that all methods achieved higher performance on NCANDA because it was the only dataset of healthy controls. Taken all together, only our approach resulted in true disentanglement between anatomical and modality representations, which was not guaranteed by the baselines.

\subsection{Evaluation on downstream tasks}
% ZeroDose
% figure of reconstructed PET
% zero-dose PET figure (w/o missing)
\begin{figure}[!t]
\centering
\includegraphics[width=\textwidth]{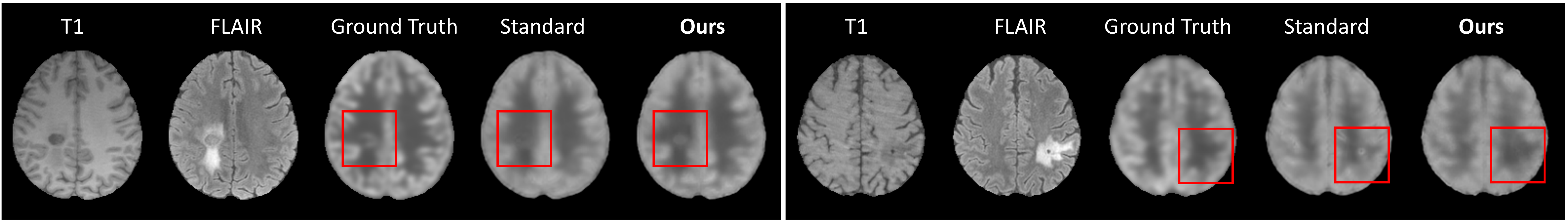}
\caption{ZeroDose PET reconstruction from T1 and FLAIR for two test subjects.} 
\label{fig:stage2-zerodose}
\end{figure}

Deep learning models that rely on multi-modal input often suffer from the missing input problem. When a model is trained on data with complete inputs and tested on data with one modality missing, standard approaches either fill in all zero values (\textbf{Standard+Zero}) or use the average image of that modality over the entire cohort (\textbf{Standard+Avg}). Here, we demonstrate that an alternative solution is to train and test the model on the fusion of disentangled anatomical representations from all available modalities (\textbf{Ours}, CondConv+Sim). We show that this strategy can largely alleviate the impact of missing inputs.

%a potentially better solution is by training models   the proposed method could further solve the incomplete inputs problem in downstream tasks. We compare with methods \textbf{Standard+Zero} that uses all zero instead if the modality is missing and \textbf{Standard+Avg} that utilizes the average image of the cohort. Methods are compared in N/A scenario, i.e. the inputs are complete, and in missing one modality scenario.

In each run of the cross-validation, we first learned the disentangled representations and then trained the downstream models (based on the raw multi-modal images for the standard methods or fused anatomical representations for our proposed method) for zero-dose PET reconstruction. Then, the downstream model was tested on data with or without missing modalities. When all modalities were used for testing, the standard and proposed approaches achieved comparable reconstruction accuracy in terms of both PSNR and SSIM (N/A in Table \ref{tab:stage2}), but we observe that our approach could generally result in more accurate tumor reconstruction (Fig. \ref{fig:stage2-zerodose}), which could not be reflected in the whole-brain similarity measure. The improvement of our approach became evident when one of the modalities was missing (Table \ref{tab:stage2}). 
%Note, there is one modality available for these scenarios, indicating the fusion module can work with single anatomical representation. 
In particular, missing FLAIR induced larger performance drop for all approaches, as ZeroDose contained a large number of images with tumor, which was more pronounced in FLAIR than T1.

\begin{table}[!t]
\tiny
\begin{subtable}{\linewidth}
\centering
\begin{tabular}{c|c|c|c||c|c|c|c|c}
    \toprule
    \multirow{3}{*}{Methods} & \multicolumn{3}{c||}{ZeroDose} & \multicolumn{5}{c}{BraTS} \\
    \cline{2-9}
     & N/A & T1 & FLAIR & N/A & T1 & T1Gd & T2 & FLAIR\\
    \cline{2-9}
    & PSNR/SSIM  & PSNR/SSIM & PSNR/SSIM  & \multicolumn{5}{c}{DICE}\\
    \hline
    Standard+Zero & \multirow{2}{*}{\textbf{25.475}/\textbf{0.739}} & 18.122/0.436 & 18.8863/0.435 & \multirow{2}{*}{\textbf{0.826}} & 0.364 & 0.240 & 0.616 & 0.298\\
    Standard+Avg & & 24.425/0.676 & 23.137/0.631 & & 0.724 & 0.279 & 0.733 & 0.452\\
    \textbf{Ours} & 25.386/0.729 & \textbf{24.610}/\textbf{0.682} & \textbf{24.193}/\textbf{0.674} & 0.821 & \textbf{0.782} & \textbf{0.779} & \textbf{0.758} & \textbf{0.772}\\
    \bottomrule
    %\multicolumn{9}{c}{}
\end{tabular}
\end{subtable}
\vspace{5pt}
\caption{Performance of two downstream tasks with incomplete input. Left: zero-dose PET reconstruction; Right: brain tumor segmentation.}
\label{tab:stage2}
\vspace{-8pt}
\end{table}
Next, we replicated this experiment for the downstream task of brain tumor segmentation on the BraTS dataset and measured the performance using the dice coefficient (DICE; Each cell in Table \ref{tab:stage2} records the average DICE across three categories: ET, ED, and NCR/NET). In line with the results of the ZeroDose experiment, the standard and proposed methods both obtained similar DICE scores, when complete inputs were used during testing. Standard+Zero recorded the lowest accuracy when missing any modality. Standard+Avg was less impacted when T1 or T2 was missing, but more impacted by T1Gd and FLAIR as the standard model relied mostly on those two modalities for localizing the tumor. The proposed method achieved the highest DICE score in all scenarios, among which the missing T2 recorded the largest drop on DICE. This might be because T2 had the most distinct appearance compared to other modalities, thus having the largest impact on the fused representation.

\section{Conclusion}
In this paper, we first proposed a novel margin loss to regularize the within-subject across-modality similarity between representations with respect to the across-subject within-modality similarity. It alleviates the information leakage problem in existing disentanglement methods. We further introduced a modified conditional convolution layer to enable training a single model for multiple modalities. Lastly, we proposed a fusion function to combine the disentangled anatomical representations from available modalities as a set of modality-invariant features for downstream tasks. Experiments on three brain MR datasets and two downstream tasks demonstrated that the proposed method achieved meaningful and robust disentangled representations compared with the existing methods. Though we only evaluated on brain images, the method is likely to generalize to other organs as long as the assumption on the within-subject across-modality and across-subject within-modality similarity holds. 
%We will further examine the generalizability of the propose method on other organ MR images and other imaging modalities.
% this can be omitted if space becomes an issue in the end
%For future work, we will generalize the framework from multi-contrast MRIs to multi-modal data and to disentangle abnormal regions from normal anatomy.

% table of PSNR/SSIM/RMSE
% BraTS
% table of Dice/Hausdorff (mean of three label)

\noindent{\bf Acknowledgement:} %The NCANDA MRIs were part of the public data release NCANDA\_PUBLIC\_4Y\_STRUCTURAL\_V01 (DOI: 10.7303/syn22216457) \cite{ncanda}, whose collection and distribution 
This work was supported by NIH funding AA021697 and by the Stanford HAI AWS Cloud Credit. 
\bibliographystyle{splncs04}
\bibliography{mybib}

\end{document}